\def\year{2021}\relax
\documentclass[letterpaper]{article} 

\usepackage{amsfonts}
\usepackage{subfigure}

\usepackage{aaai21}  
\usepackage{times}  
\usepackage{helvet} 
\usepackage{courier}  
\usepackage[hyphens]{url}  
\usepackage{graphicx} 
\usepackage{graphicx}  
\usepackage{natbib}  
\usepackage{caption} 

\usepackage{mathrsfs} 
\usepackage{amsmath}
\title{LRC-BERT: Latent-representation Contrastive Knowledge Distillation for Natural Language Understanding}
\author{
	Hao Fu,\textsuperscript{\rm 1}\footnote{Hao Fu and Shaojun Zhou are equal contribution first authors. This work was mainly done when Hao Fu was an intern at Alibaba Group.}
	Shaojun Zhou,\textsuperscript{\rm 2}\footnotemark[\value{footnote}]
	Qihong Yang,\textsuperscript{\rm 2}
	Junjie Tang,\textsuperscript{\rm 2}\thanks{Junjie Tang is the corresponding author.}
	Guiquan Liu,\textsuperscript{\rm 1}\\
	Kaikui Liu,\textsuperscript{\rm 2}
	Xiaolong Li\textsuperscript{\rm 2}
	\\
}
\affiliations{
	\textsuperscript{\rm 1}School of Computer Science and Technology, University of Science and Technology of China\\
	\textsuperscript{\rm 2}Alibaba Group\\
	
	hfu@mail.ustc.edu.cn, \{zsj148798, xiaokui.yqh, lixi.tjj\}@alibaba-inc.com, gqliu@ustc.edu.cn,\\
	\{damon, xl.li\}@alibaba-inc.com
	
}

\begin{document}

\maketitle

\begin{abstract}

The pre-training models such as BERT have achieved great results in various natural language processing problems. However, a large number of parameters need significant amounts of memory and the consumption of inference time, which makes it difficult to deploy them on edge devices. In this work, we propose a knowledge distillation method LRC-BERT based on contrastive learning to fit the output of the intermediate layer from the angular distance aspect, which is not considered by the existing distillation methods. Furthermore, we introduce a gradient perturbation-based training architecture in the training phase to increase the robustness of LRC-BERT, which is the first attempt in knowledge distillation. Additionally, in order to better capture the distribution characteristics of the intermediate layer, we design a two-stage training method for the total distillation loss. Finally, by verifying 8 datasets on the General Language Understanding Evaluation (GLUE) benchmark, the performance of the proposed LRC-BERT exceeds the existing state-of-the-art methods, which proves the effectiveness of our method.

\end{abstract}

\section{Introduction}

Recently, one of the main trends of natural language processing (NLP) is pre-training model~\cite{peters2018deep,radford2018improving}. The pre-training models such as BERT~\cite{devlin2019bert} and XLNet~\cite{yang2019xlnet} have achieved remarkable results in  tasks like Sentiment Classification~\cite{socher2013recursive} and Natural Language Inference~\cite{williams2017broad}. The training of these models usually consists of two stages: the first stage is the model pre-training on a large scale corpus by predicting specific words according to the given context. The second stage is to add a specific prediction layer on a specific downstream task for fine-tuning training. However, these models usually contain hundreds of millions of parameters. For example, the original BERT-base~\cite{devlin2019bert} model has 12 layers and 109 million parameters. This limits the deployment of online services because of inference delay and device capacity constraints. Therefore, in the case of limited computing resources, it is necessary to reduce the computational cost of these models in practice.

Knowledge distillation (KD)~\cite{hinton2015distilling,adriana2015fitnets} is an effective method of model compression~\cite{han2015deep}. By compressing a large teacher network into a small student network, the student network has the same predictive ability as the teacher network. At the same time, the student network has fewer parameters, correspondingly faster inference speed on specific tasks and can be effectively deployed to edge devices, greatly saving computing resources. Recalling the existing research on the KD method in NLP, DistilBERT~\cite{sanh2019distilbert} uses soft label distillation loss, cosine embedding loss, and initializes the student from the teacher by taking one layer out of two. This strategy forces the structure of the transformer layer in the student network to be consistent with the teacher, and the distillation of the output characteristics of the intermediate layer is not fully considered. The student model in BERT-PKD~\cite{sun2019patient} patiently learns from multiple intermediate layers of the teacher model for incremental knowledge extraction, including PKD Last and PKD Skip. The former try to learn from the last $k$ layers, the latter learn from each $k$ layer. Like DistilBERT, BERT-PKD also requires the corresponding layer in the student to be exactly the same as the teacher, which greatly limits the flexibility of structural design and the compression scale. TinyBERT~\cite{jiao2019tinybert} uses more fine-grained knowledge including the hidden state of the transformer network and self-attention distribution. Additionally, a parameter matrix is introduced to perform a linear transformation on the hidden state of student so as to reduce the number of parameters in the tranformer layer of the student. However, these models only mimic the output value of each layer of the teacher network, without considering structural information such as the correlation and difference of the output distribution among different samples.

In this work, we design a model compression framework called LRC-BERT to distill the knowledge of the teacher network into the student network. Specifically, we design a variety of different loss functions for the output of the transformer and prediction layer. For the output logits of the prediction layer, we calculate Kullback–Leibler (KL) divergence and cross entropy with the output of the teacher and the real label respectively; for the output of the intermediate transformer layer, we design a novel contrastive loss cosine-based noise-contrastive estimation (COS-NCE) to capture the distribution structure characteristics of the output of the intermediate layer. The design of COS-NCE is inspired by infoNCE~\cite{oord2018representation} and CRD~\cite{DBLP:conf/iclr/TianKI20}. Optimizing the contrastive loss can narrow the gap between the representations of positive pairs in a metric space, push the gap between the representations of negative pairs away, and learn the structural information output by the teacher network in the intermediate layer effectively. However, these structural information is ignored in the existing methods, such as TinyBERT and MINILM~\cite{wang2020minilm}, they only approximate the output values of the teacher network in the transformer layers. Compared with existing model DistilBERT, LRC-BERT choose a more flexible student structure without the limitation of the structure and parameter quantity of the intermediate layer to be consistent with the teacher by introducing a dimension transformation matrix. Then, in order to help LRC-BERT focus on the structural information of the output distribution in the early stage of training, we adopt a two-stage training method for the total distillation loss, and verify the effectiveness of calculating distillation loss by stages in the experimental part. Finally, we introduce a new model training architecture based on gradient perturbation~\cite{DBLP:conf/iclr/MiyatoDG17}, which perturbs the output of the word vector embedding layer through the gradient value, thereby changing the output distribution of the model in this layer and improving the robustness of the model. Our contributions are summarized as follows:

\begin{quote}
\begin{itemize}
\item For the distillation of the intermediate layer, a new contrastive loss COS-NCE is proposed to effectively capture the structural characteristics between different samples.
\item The gradient perturbation is introduced for the first time in knowledge distillation and verified in the experimental part that it can improve the robustness of LRC-BERT.
\item We use a two-stage training method to better capture the output distribution characteristics of the intermediate layer based on COS-NCE.

\item  We evaluate the proposed method on 8 NLP datasets, and the experimental results show that our LRC-BERT model outperforms than the state-of-the-art baseline models.
\end{itemize}
\end{quote}

\section{Related Work}

\subsection{NLP Pre-trained Language Models}

The emergence of pre-training models has brought NLP into a new era~\cite{peters2018deep,song2019mass,jiao2019tinybert,devlin2019bert}. The advantages of using pre-training models are as follows: 1) Pre-training models learn knowledge from the large-scale corpus, which is helpful for downstream tasks; 2) pre-training provides a better parameter initialization method, which makes better generalization and faster convergence on the target task; 3) pre-training can be considered as a regularization method, which can prevent the model from over fitting on small datasets~\cite{qiu2020pre}. Specifically, the pre-training language model can be divided into two stages: 1) pre-training word embedding; 2) pre-training context encoder.

In the first stage of pre-training word embedding, only word vectors are trained, which is a static feature-based mode. Typical examples are Word2Vec~\cite{mikolov2013efficient} and Glove~\cite{pennington2014glove}. The word vector is used for token embedding and sent it into the specific model. This kind of model has simple structure, but it can also get high-quality word vector which can capture the potential grammatical and semantic information between words in the potential text. However, this type of pre-training word vector can not be dynamic with the context variety. In addition, in the downstream tasks, the remaining model parameters still need to be retrained.

In the second stage of pre-training context encoder, the semantic information dynamically changes with the context. Typical examples are ULMFiT~\cite{howard2018universal}, ELMo~\cite{peters2018deep} and GPT~\cite{radford2018improving}. With the introduction of transformer~\cite{vaswani2017attention}, it has brought more attention to the pre-training language model with a deeper structure. The typical representative is BERT~\cite{devlin2019bert}, which is a bi-directional encoder based on transformer and can learn sentence information more completely by using context. Through pre-training via masked language modeling and next sentence prediction, it has achieved the advanced performance on many NLP tasks, such as the GLUE benchmark~\cite{wang2019glue}.

\subsection{Contrastive Learning}

Contrastive Learning~\cite{hadsell2006dimensionality,weinberger2009distance,schroff2015facenet}, as a kind of self-supervised learning method, uses the data itself as the supervised information to learn the feature expression of the sample data by constructing positive samples and negative samples, without manual annotation information. In fact, by minimizing some specially constructed contrastive loss such as infoNCE~\cite{oord2018representation} and CRD~\cite{tian2019contrastive}, the lower bound of the mutual information between all samples can be maximized.

The construction of positive and negative samples is one of the major difficulties in contrastive learning. In the field of image, deep InfoMAX~\cite{hjelm2018learning} takes local features of training images and different images as positive samples and negative samples respectively. Then, MoCo~\cite{he2020momentum} and SimCLR~\cite{chen2020simple} extract positive samples by clipping, rotation and other transformation operations. For the data such as text and audio, because the data itself is sequence, the method adopted by the CPC~\cite{oord2018representation} is to take the sample from the data that has not been input into the model as a positive example, and randomly sampling an input from the input sequence as a negative example.

\subsection{Knowledge Distillation}
The model compression technology~\cite{han2015deep} accelerates the inference speed and reduces the number of model parameters while ensuring the prediction performance of the model. Its main technologies consist of using fewer bits to represent the quantization of parameter weights~\cite{gong2014compressing}, weight pruning~\cite{han2015learning} to reduce or dilute network connections and knowledge distillation~\cite{adriana2015fitnets,tang2019distilling,tan2018multilingual}.

Knowledge distillation has proven to be a promising method in model compression, which transfers the knowledge of a large model or a group of neural networks (teacher) to a single lightweight model (student). ~\cite{hinton2015distilling} first proposed using the soft target distribution to train the student model and impart the knowledge of teachers to students.~\cite{DBLP:journals/corr/ChenGS15} introduced technologies that effectively transfer knowledge from existing networks to a deeper or wider network.~\cite{turc2019well} used a small pre-trained language model to initialize students during task-specific distillation. MiniBERT~\cite{tsai2019small} uses soft target distribution for mask language modeling prediction, guides the training of multilingual student models. TinyBERT~\cite{jiao2019tinybert} and MobileBERT~\cite{DBLP:conf/acl/SunYSLYZ20} further introduced self-attention distribution and hidden state to train students.~\cite{aguilar2020knowledge} distilled internal representation knowledge from attention layer and intermediate output layer in a bottom-up way.

Different from the existing works paying more attention to approximate the attention value and output value in the transformer layer, we try to capture the correlation and difference of the output distribution between different input data of the teacher network in each layer, and these structural features have also been proved to be feasible in the knowledge distillation by CRD.

\begin{figure}[t]
	\centering
	\includegraphics[width=0.75\columnwidth]{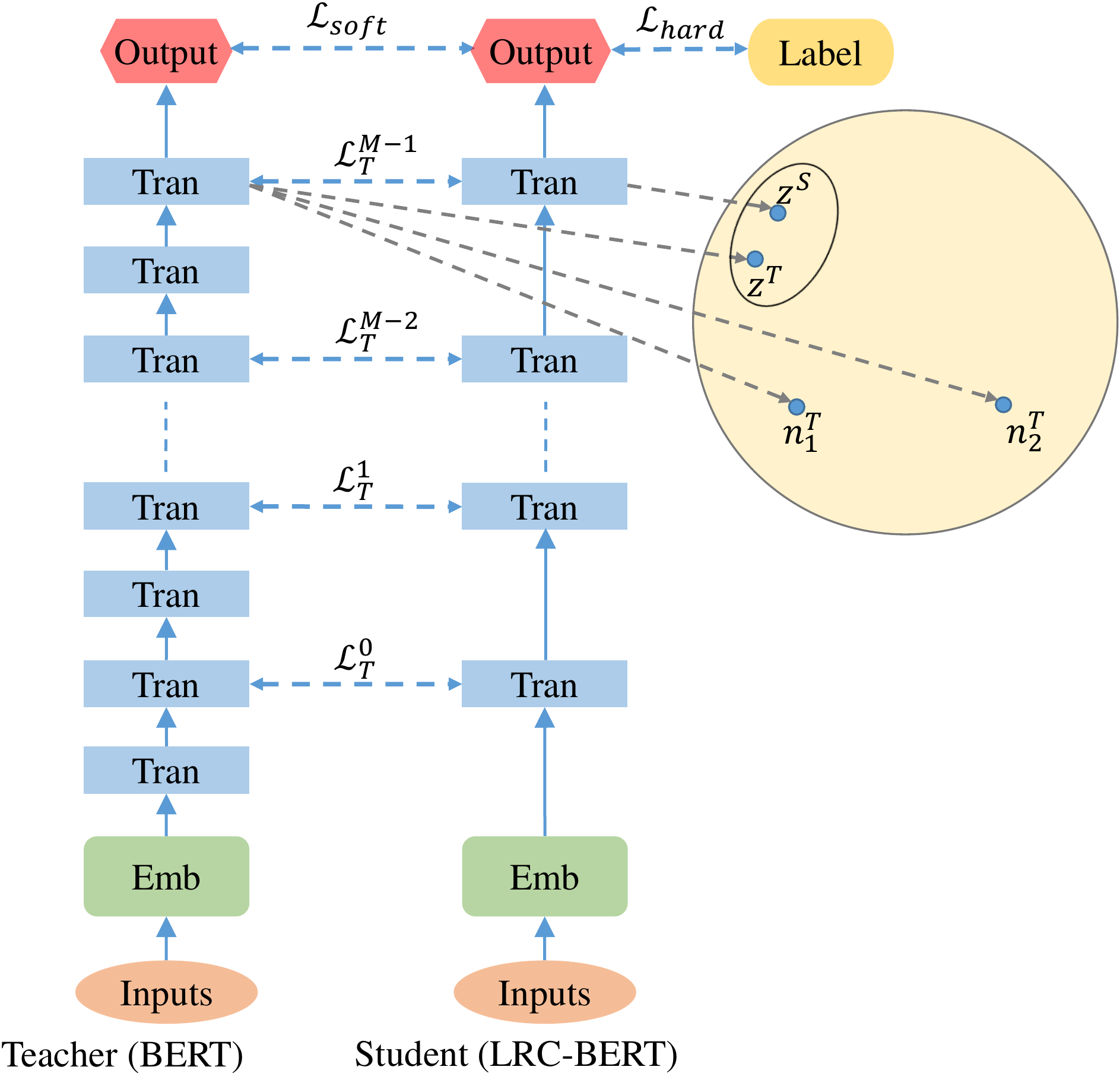} 
	\caption{Overview of the proposed latent-presentation contrastive distillation framework for BERT model compression. Among them, contrastive learning is used to transfer the structural characteristics of output distribution from the teacher to the student. For instance, for the same input sample, the output distribution $z^S$ of the student is more similar to $z^T$ that of the teacher when comparing with the outputs $n_1^T$ and $n_2^T$ of the teacher for the negative input samples.}
	\label{fig_structure}
\end{figure}

\section{Method}

In this section, we first propose COS-NCE by distilling the output distribution of the intermediate layer, and then show the process details of proposed adaptive deep contrastive distillation framework LRC-BERT, as shown in Fig.~\ref{fig_structure}. Finally, we introduce a model training method based on gradient perturbation to enhance the robustness of LRC-BERT.

\subsection{Problem Definition}
The trained teacher network is represented by the function $f^T(x,\theta)$, where $x$ is the input of the network and $\theta$ is the parameters. Our goal is to make the student network $f^S(x,\theta')$ imitate the output distribution of $f^T(x,\theta)$ in the intermediate and prediction layer by minimizing the specific loss $\mathcal{L}_{total}$, so as to realize the purpose of knowledge distillation.

\subsection{COS-based NCE loss}

\begin{figure}
	\centering
	
	\subfigure[]{
		\begin{minipage}[t]{0.19\textwidth}
			\centering
			\includegraphics[width=\textwidth]{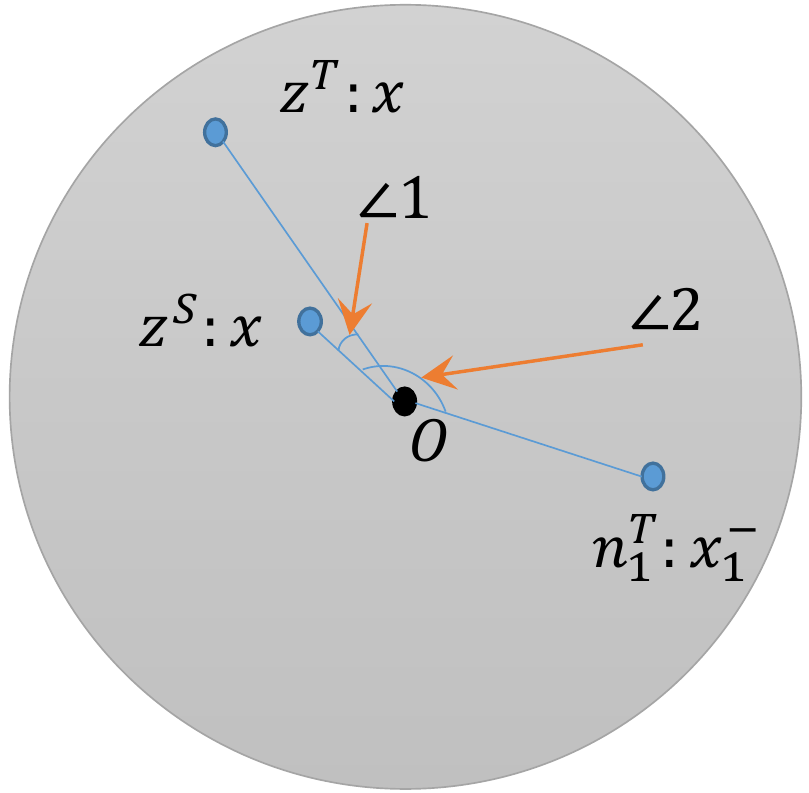}
			
		\end{minipage}%
	}
	\subfigure[]{
		\begin{minipage}[t]{0.19\textwidth}
			\centering
			\includegraphics[width=\textwidth]{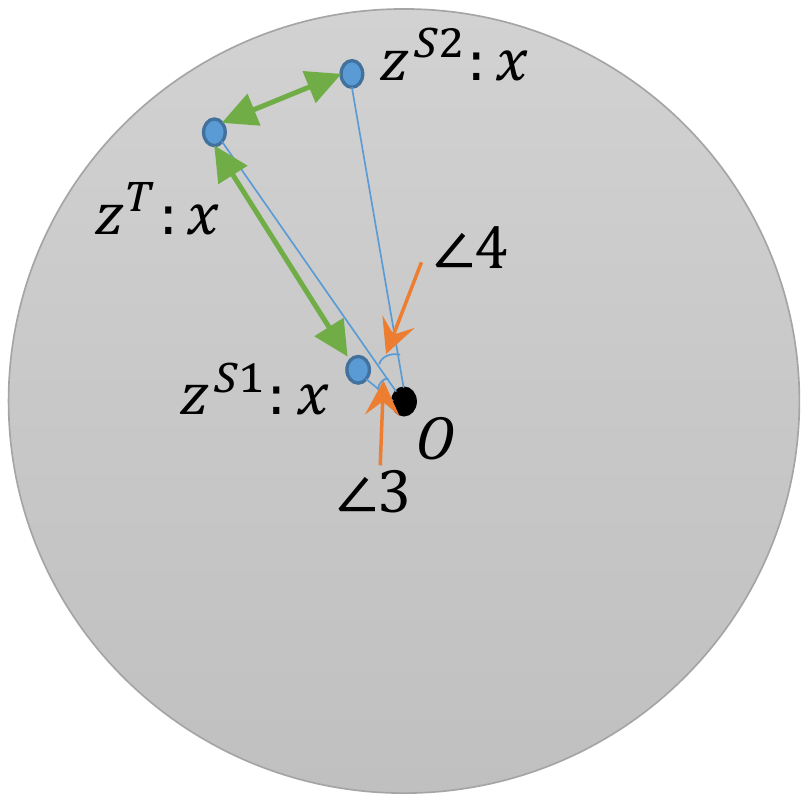}
			
		\end{minipage}%
	
	}%

	\caption{Map the output of the teacher network and the student model in the penultimate layer to the same feature space. The green line represents Euclidean distance. (a) the angular distance between positive pairs $z^S$ and $z^T$ with the same input $x$ is smaller than that between negative pairs $z^S$ and $n^T_1$ with different inputs $x$ and $x_1^-$, (b) for different student models $f^{S1}$ and $f^{S2}$, the distance between the output $z^{S2}$ and $z^T$ is closer than that of $z^{S1}$ under the evaluation standard based on Euclidean distance; however, under the standard based on angular distance, $f^{S1}$ is better than $f^{S2}$.}\label{fig_1}
\end{figure}
\begin{figure}[t]
	\centering
	\includegraphics[width=0.9\columnwidth]{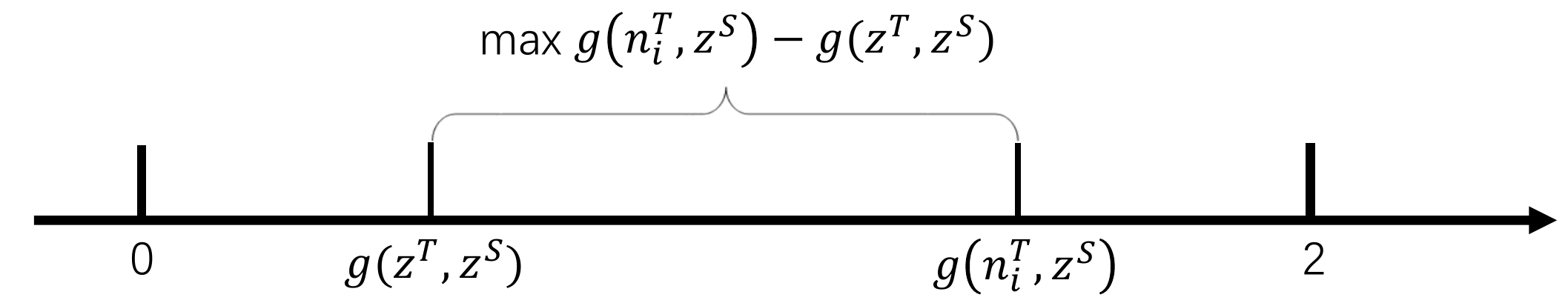} 
	\caption{We use the difference between $g(n_i^T,z^S)$ (similarity with negative sample) and $g(z^T,z^S)$ (similarity with positive sample) to represent the distance from the negative sample. }
	\label{fig_3}
\end{figure}

Inspired by recent contrastive learning algorithms~\cite{oord2018representation,tian2019contrastive,sohn2016improved}, we design a contrastive loss COS-NCE for the transfer of the intermediate layer. In particular, we use the penultimate layer of distillation to elicit our proposed COS-NCE. For a given teacher network $f^T$ and student model $f^S$, we randomly sample K negative samples $X=\{x_1^-,x_2^-,...,x_K^-\}$ for each input training sample $x$, and record the output of sample $x$ in the penultimate layer of $f^T$ and $f^S$ networks as $z^T$ and $z^S$, respectively. Correspondingly, we can obtain the output $N=\{n^T_1,n^T_2,...,n^T_K\}$ of negative samples in the penultimate layer of $f^T$ network. As shown in Fig.~\ref{fig_1}, we map the outputs of the penultimate layer to certain points on the high-dimensional feature space. Different from the previous contrastive loss works considering Euclidean distance or mutual information in feature space~\cite{sun2014deep}, we propose a new contrastive loss COS-NCE based on angular distance distribution to consider the latent structural information output by the intermediate layer.

Formally, distilling the output of a given sample and K negative samples in the penultimate layer, the proposed COS-NCE formula is as follows:

\begin{gather}
\begin{split}\label{formula1}
\mathcal{L}_C(z^S,z^T,N)=&\frac{\displaystyle\sum_{i=1}^{K}(2-(g(n^T_i,z^S)-g(z^T,z^S))}{2K}\\
&+g(z^T,z^S),\\
\end{split}
\end{gather}%
\begin{equation}
g(x,y)=1-\frac{x\cdot y}{\Vert x\Vert\Vert y\Vert},\label{eq2}
\end{equation}
where $g(.,.)\to [0,2]$ is a function used to measure the angular distance between two input vectors, and the smaller $g(x,y)$ means that the distribution distance of the two vectors is closer to each other. When $g(x,y) = 2$, it means that the two vectors reach the maximum dissimilarity. The purpose of COS-NCE in formula~\eqref{formula1} is to minimize the angle between $z^S$ and the positive sample $z^T$, while increasing the angle between $z^S$ and the negative sample $n^T_i\in N$. Instead of constructing the form of triplet loss based on Euclidean distance as used in previous work~\cite{weinberger2009distance,schroff2015facenet}, we apply the function $g (.,.)$ to minimize the angular distance between the training sample and positive samples. Then, in order to maximize the distance between the training sample and the negative samples (the first term in formula~\eqref{formula1}), the loss is constructed from the geometric perspective as shown in Fig.~\ref{fig_3}. Obviously, the difference value $g(n^T_i,z^S)-g(z^T,z^S)$ between $g(n^T_i,z^S)$ and $g(z^T,z^S)$ need to be maximized. In order to transform the maximization problem into the minimization equivalent problem, we calculate the distance from $g(z^T,z^S)$ to the lower boundary of the function and $g(n^T_i,z^S)$ to the upper boundary. Therefore, the maximization problem is redefined as minimizing the sum of these two distances: $2-(g(n^T_i,z^S)-g(z^T,z^S))$.

\subsection{Distillation for transformer-layer}

\begin{table*}  
	\begin{small}

	\centering
	\begin{tabular}{c|c|ccccccccc|c} 
		\hline  
		Model & Params & MNLI-m  & MNLI-mm& QQP& SST-2& QNLI& MRPC& RTE& CoLA& STS-B &Avg \\  
		& & (393k)  & (393k)& (364k)& (67k)& (105k)& (3.7k)& (2.5k)& (8.5k)& (5.7k) & \\  
		\hline  
		BERT-base (teacher) & 109M   &  84.3 &83.8 &71.4 &93.6 &90.9 &88.0 &66.4 &53.0 &84.8  &79.6\\ 
		\hline  
		
		DistilBERT & 52.2M & 78.9  & 78.0& 68.5& 91.4& 85.2& 82.4& 54.1& 32.8& 76.1 &71.9\\ 
		BERT-PKD &  52.2M  & 79.9  & 79.3&70.2 &89.4 &85.1 &82.6 &62.3 &24.8 &79.8  &72.6\\ 
		TinyBERT &  14.5M  & 82.5  & 81.8& 71.3& 92.6& 87.7& 86.4& 62.9& 43.3& 79.9 &76.5\\ 
		LRC-BERT$_1$ &14.5M &  82.8 &82.6 &71.9 &90.7 &88.3 &83.0 &51.0 &31.6 & 79.8 &73.5\\ 
		LRC-BERT  & 14.5M  &  \textbf{83.1} & \textbf{82.7}& \textbf{72.2}& \textbf{92.9}& \textbf{88.7}& \textbf{87.0}& \textbf{63.1}& \textbf{46.5}& \textbf{81.2} &\textbf{77.5}\\ 
		\hline  
	\end{tabular}  
\end{small}
	\caption{The results are evaluated from the official website of GLUE benchmark, and the optimal experimental results are identified in bold. The number under each dataset represents the corresponding number of training samples.}\label{table-1}
\end{table*}

The COS-NCE is proposed to transfer the output of transformer layer knowledge from the teacher network to the student model. Each transformer layer in BERT consists of two main sub-layers: multi head attention and fully connected feed-forward network (FFN), but we only distill the output of FFN.  Assuming that the teacher model has $N$ transformer layers and the student model has $M$ transformer layers, we need to select $M$ layers in the teacher for knowledge transfer and the objective is as follows:

\begin{equation}
\mathcal{L}_T^i=\mathcal{L}_C(h^S_i W,h^T_{\phi_{i}},H^T_{\phi_{i}}),
\end{equation}
where $h^S_i \in \mathbb{R}^{l\times d}$ is the output of the training sample $x$ in the $i$-th transformer layer of the student network, $h^T_{\phi_{i}} \in \mathbb{R}^{l\times d'}$ is the output of the training sample $x$ in the ${\phi_{i}}$-th transformer layer of the teacher model, ${j=\phi_{i}}$ is a mapping function, which means that the output of the $i$-th layer of student network needs to imitate the output of the $j$-th layer of teacher model, $l$ is the input text length, and scalar values $d$ and $d'$ represent the hidden size of student and teacher respectively, usually $d$ is smaller than $d'$. $H^T_{\phi_i}=\{h^T_{0,\phi_i},h^T_{1,\phi_i},...,h^T_{K-1,\phi_i}\}$ is the output of K negative samples in the $i$-th transformer layer of teacher network. Referring to the exiting works ~\cite{jiao2019tinybert,wang2019heterogeneous}, we use matrix $W\in \mathbb{R}^{d\times d'}$ to map the outputs of different dimensions to the same feature space.

\subsection{Distillation for predict-layer}

In order to predict specific downstream classification tasks, we simulate the student's prediction output from the soft label of the teacher model and the hard label corresponding to the real label respectively. Among them, the KL divergence is used to imitate the output of the teacher network, and cross-entropy loss is used to fit the difference between the student's output and the one-hot encoding of the real label.

\begin{gather}
\mathcal{L}_{soft}=\text{softmax}(y^T/\tau)\cdot \text{log}(\frac{\text{softmax}(y^T/\tau)}{\text{softmax}(y^S/\tau)}),\\
\mathcal{L}_{hard}=-\text{softmax}(y/\tau)\cdot\text{log}(\text{softmax}(y^S)),
\end{gather}
where $y^S$ and $y^T$ are logits vectors output by the predict layer of student and teacher respectively, $\tau$ is a temperature that adjusts the concentration level, $y$ is the $one-hot$ encoding of real label. It should be noted that there is no negative sample involved in the distillation of the prediction layer. When the downstream task is a regression problem, the calculation of $\mathcal{L}_{soft}$ and $\mathcal{L}_{hard}$ is changed to mean squared error (MSE) loss function.

The various losses described above are shown in Fig~\ref{fig_structure}. By combining the various losses of different layers described above, we can obtain the final distillation objective function for transforming the knowledge of the teacher network into the student model:
\begin{equation}\label{formula6}
\mathcal{L}_{total}=\alpha\sum_{i=0}^{M-1}\mathcal{L}_T^i+\beta \mathcal{L}_{sotf}+\gamma \mathcal{L}_{hard},
\end{equation}
where $\alpha$, $\beta$ and $\gamma$ are used to weigh the influence of different losses. Specifically, in order to enable the model to learn the contrastive structural features more effectively in the preliminary training, we will adopt a two-stage training method. In the first stage, we only calculate the contrastive loss of the intermediate transformer layer, that is, the values of $\alpha$, $\beta$ and $\gamma$ are set to 1, 0 and 0 respectively. In the second stage, the weight values are set to be greater than 0 to ensure the model has the ability to predict downstream tasks.

\begin{figure}[t]
	\centering
	\includegraphics[width=0.35\columnwidth]{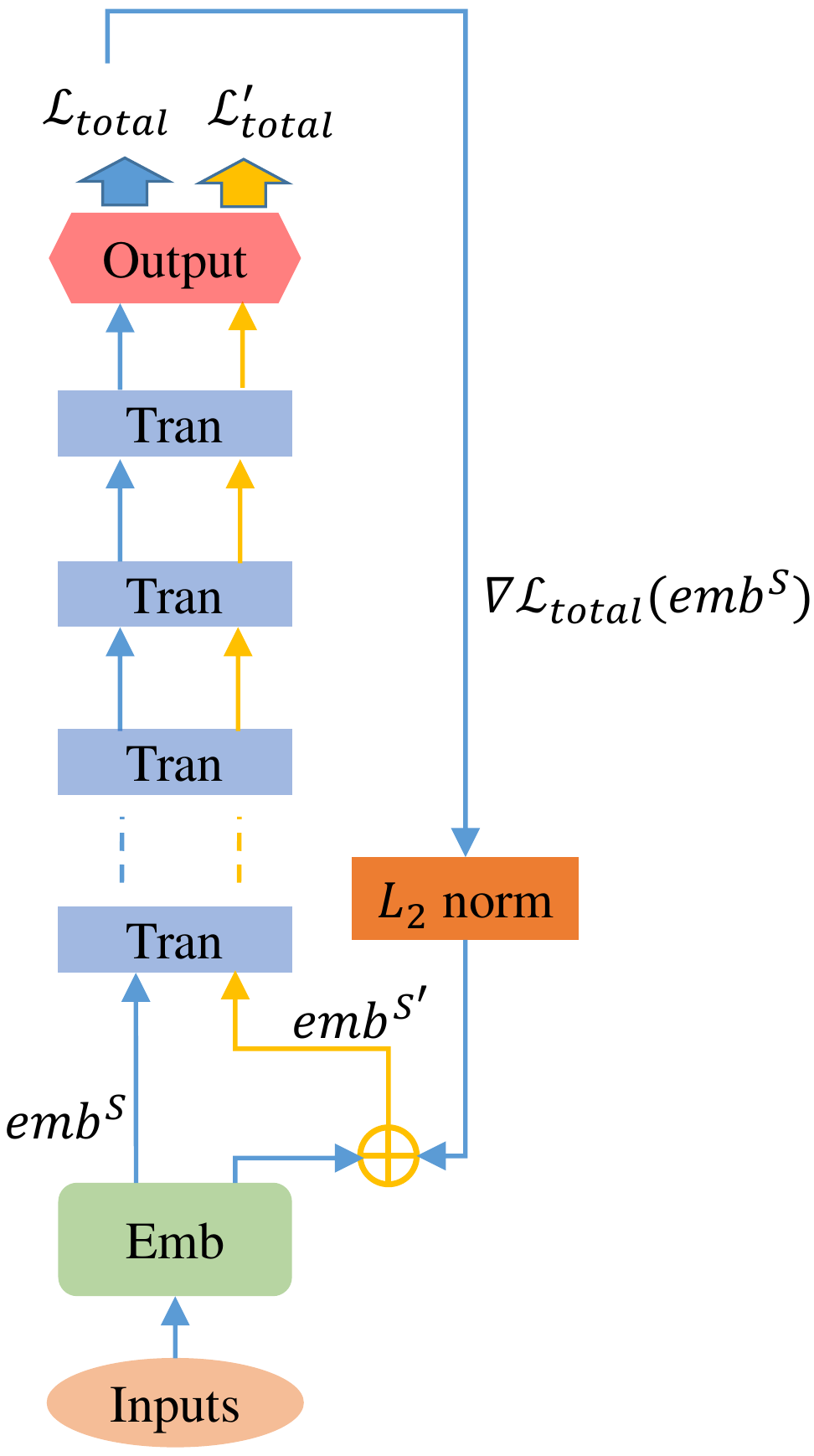} 
	\caption{The illustration of training based on gradient perturbation.}
	\label{fig_4}
\end{figure}

\subsection{Training based on Gradient Perturbation}

In the training stage, we introduce gradient perturbation to increase the robustness of LRC-BERT. The method based on gradient perturbation has been applied in word embeddings~\cite{DBLP:conf/iclr/MiyatoDG17}, and the prediction performance is more stable. Fig.~\ref{fig_4} shows the training process based on gradient perturbation. For the distillation loss $\mathcal{L}_{total}$ obtained according to formula~\eqref{formula6}, we do not directly use the loss to carry out the parameter update by backward gradient propagation, but calculate the gradient of each element in the output matrix $emb^S$ of the embedding layer in the student network, and use the gradient value to interfere with the distribution of $emb^S$. Then use the distribution after interference to calculate the loss according to the network forward propagation, and the loss is finally used to update the network parameters. The specific process is as follows:

\begin{equation}
	emb^{S'}=emb^S+g/||g||_2,\\
\end{equation}
\begin{equation}
g=\nabla\mathcal{L}_{total}(emb^S).
\end{equation}

Where $emb^{S'}$ is the vector after using gradient perturbation to interfere, and the calculation of the gradient perturbation is $L_2$ norm processing for the gradient value $emb^S$. Before the parameter of the student network is updated, it is passed into the subsequent transformer layer and prediction layer again to calculate the loss $\mathcal{L}_{total}’$ after the gradient perturbation.

\section{Experiment}
 
In this section, we elaborate on a large number of experiments to verify the effectiveness of the proposed LRC-BERT. The specific details are shown in the following subsections.
\subsection{Datasets}
We evaluate LRC-BERT on GLUE benchmark. The datasets provided on GLUE were all from NLP datasets with high recognition. We evaluate LRC-BERT in tasks such as natural language reasoning, emotion analysis, reading comprehension and semantic similarity.

In more detail, we evaluate on 8 classification tasks, including Corpus of Linguistic Acceptability (CoLA)~\cite{warstadt2019neural}, Stanford Sentiment Treebank (SST-2)~\cite{socher2013recursive}, Microsoft Research Paraphrase Corpus (MRPC)~\cite{dolan2005automatically}, Semantic Textual Similarity Benchmark (STS-B)~\cite{cer2017semeval}, Quora Question Pairs (QQP)~\cite{chenquora}, Question Natural Language Inference (QNLI)~\cite{rajpurkar2016squad}, Recognizing Textual Entailment (RTE)~\cite{bentivogli2009fifth} and Multi-Genre Natural Language Inference (MNLI)~\cite{williams2017broad}. Among them, MNLI is further divided into two parts: in-domain (MNLI-m) and cross-domain (MNLI-mm), so as to evaluate the generality of the test model. When evaluating the model, we use the metrics used by GLUE benchmark. Specifically, For QQP and MRPC, the metric is F1-score, for STS-B, the metric is Matthew's correlation, the rest tasks use accuracy.

\subsection{Distillation Setup}
We use BERT-base~\cite{devlin2019bert} as our teacher. BERT-base contains a total of about 109M parameters, including 12 layers of transformer ($N=12$), and each layer has 768 hidden dimensions ($d'=768$), 3072 intermediate sizes and 12 attention heads. The student model we instantiated has 4 transformer layers ($M=4$), 312 hidden sizes ($d=312$), 1200 intermediate sizes and 12 attention heads, with a parameter of approximately 14.5M. In order to test the generality of our model more effectively, we design two groups of experimental procedures. The first group uses the Wikipedia corpus to conduct pre-training, and then distills on specific tasks, which is called LRC-BERT. In the pre-training stage, only one epoch of distillation is carried out on the transformer layers, the values of $\alpha$, $\beta$ and $\gamma$ are set to 1, 0 and 0 respectively. The second group is directly distilled under the specific task dataset and recorded as LRC-BERT$_1$.

For the distillation of each task on GLUE, we fine-tune a BERT-base teacher, choosing learning rates of 5e-5, 1e-4, and 3e-4 with batchsize of 16 to distill LRC-BERT and LRC-BERT$_1$. For each sample, we choose the remaining 15 samples in batchsize as negative samples, i.e. $K=15$. Among them, 90 epochs of distillation are performed on the MRPC, RTE, and CoLA with the training dataset less than 10K, and 18 epochs of distillation on other tasks. For the proposed two-stage training method, the first 80\% of the steps are chosen as the first stage of training, the rest 20\% of the steps are the second stage. Specifically, the partition parameter value 80\% applies to all tasks is obtained by searching on MNLI-m and MNLI-mm tasks, and the search range is \{0\%, 5\%, 10\% ,...,95\%\}. Then, we set the parameters of the second stage to $\alpha: \beta: \gamma = 1:1:3$, and the search range of each parameter is \{1,2,3,4\}. For the hyperparametric temperature $\tau$, we set it to 1.1. Considering that there are fewer training samples for RTE and CoLA, we use the data augmentation method used by TinyBERT in LRC-BERT to expand the training data by 20 times and conduct 20 epochs of transformer layer distillation on the augmented dataset before specific task distillation.

We distill our student model with 6 V100 in the pre-training stage, and 4 V100 for distillation training on specific task dataset and extended dataset. In the inference experiments, we report the results of the student on a single V100.

\begin{table}  
	\begin{small}
	
	\centering
	\begin{tabular}{ccccc} 
		\hline  
		Model & transformer  & hidden& Params& inference\\  
		&  layers & size& & time(s)\\  
		\hline  
		BERT-base&  12 &768 &109M & 121.4\\ 
		LRC-BERT     &  4 & 312& 14.5M& 12.7\\ 
		\hline  
	\end{tabular}  
\end{small}
	\caption{The number of parameters and inference time before and after model compression.}  \label{table-2}
\end{table} 

\subsection{Main Results}
The results of the evaluation on GLUE are recorded in Table~\ref{table-1}. By comparing with BERT-PKD~\cite{sun2019patient}, DistilBERT~\cite{sanh2019distilbert} and TinyBERT~\cite{jiao2019tinybert}, which also have 4 transformer layers, the following three conclusions can be drawn: 1) in all evaluation tasks, LRC-BERT has better distillation effect than other methods, and the average prediction effect of LRC-BERT can reach 97.4\% of the teacher model BERT-BASE, which indicates that the proposed method is effective for distillation on specific tasks. 2) For tasks with a large amount of training data, LRC-BERT${_1}$, which does not use data augmentation and pre-training, can significantly simplify the distillation process while still effectively transferring teacher knowledge to students. Specifically, for tasks with a training set greater than 100K LRC-BERT$_1$ has surpassed the experimental effect of TinyBERT, increasing by 0.3\%, 0.8\%, 0.6\%, and 0.6\% in MNLI-m, MNLI-mm, QQP and QNLI respectively. 3) Compared with the experimental results of LRC-BERT and LRC-BERT$_1$, the distillation performance of LRC-BERT can be significantly improved by pre-training and data augmentation, especially when the amount of training data is small. For example, the prediction results on tasks MRPC, RTE and CoLA are 4.0\%, 12.1\% and 14.9\% higher than that of LRC-BERT$_1$, respectively.

Inference time is another important evaluation metric for model compression. Therefore, we further investigated the increase in model inference speed, and summarized the inference time of different baseline models in Table~\ref{table-2}. Our 4-layer 312-dimensional student network is 9.6$\times$ faster than the original teacher model BERT-BASE with the model size is 7.5$\times$ smaller. This result is similar to TinyBERT, because our student model structure is the same as TinyBERT.

\begin{table}  
\begin{small}
	\centering
	\begin{tabular}{ccccc} 
		\hline  
		Model & MNLI-m  & MNLI-mm& MRPC& CoLA\\  
		\hline  
		LRC-BERT&  83.4 &83.5 &89.0 & 50.0\\ 
		\hline  
		LRC-BERT$_C$     &  78.0 & 78.2&81.5 & 37.0\\ 
		LRC-BERT$_S$  &  82.7 &83.0 & 89.4& 48.8\\ 
		LRC-BERT$_H$     & 83.0  &83.5 &88.7 & 48.6\\ 
		\hline  
	\end{tabular} 
\end{small} 
	\caption{Ablation studies of different loss functions (dev).}\label{table-3} 
\end{table}  

\begin{table}  
	
	\centering
	\begin{tabular}{cc} 
		\hline  
		Model & Accuracy  \\  
		\hline  
		LRC-BERT     &   83.4 \\ 
		LRC-BERT$_2$  &  79.4 \\ 
		\hline  
	\end{tabular}  
	\caption{Effect of two-stage training method on MNLI-m task (dev).} \label{table-4}
\end{table} 

\begin{table*}  
	
	\centering
	\begin{tabular}{cccccc} 
		\hline  
		sentence1 & sentence2  & MSE& angular distance($g$)& label&prediction (BERT$_M$)\\  
		\hline  
		Paper goods.& Paper products.  &1.532 & 0.505& entailment&entailment\\ 
		oh constantly&  Rarely  &1.557 & 0.525& contradiction&contradiction\\ 
		The truth?&  Will you tell the truth?  &1.730 & 0.496& neutral&contradiction\\ 
		I'm not exactly sure.&  I'm not exactly sure if you're  &2.108 & 0.484& entailment&neutral\\ 
		&  aware of your issues. & & & &\\ 
		\hline  
	\end{tabular}  
	\caption{Display of prediction results: In the given 4 cases, LRC-BERT can make accurate prediction, while BERT$_M$ performs poorly, indicating that the angle-based distillation effect is better than the Euclidean distance.} \label{table-5}
\end{table*} 

\subsection{Ablation Studies}

In this section, we conduct ablation tests on MNLI-m, MNLI-mm, MRPC and CoLA from two aspects: the construction of different loss functions and the use of gradient perturbation.
\subsubsection{Effect of different loss function}

We remove the COS-NCE of transformer layer, soft loss or hard loss of prediction layer from the loss function for distillation shown in formula~\eqref{formula1} to get three contrast models of LRC-BERT$_C$, LRC-BERT$_S$ and LRC-BERT$_H$. The evaluation results of these four tasks on dev are shown in Table~\ref{table-3}. The experimental results indicate that the COS-NCE plays the most important role in distillation. When COS-NCE is discarded, the performances are getting worse in all datasets, especially on the CoLA task, which is drop from 50 to 37. In addition, the absence of soft-loss or hard-loss shows little effect on the results. It is worty to noted that they have different degrees of influence on different tasks. For example, the prediction performance of soft loss used in MRPC and CoLA are outperform than that of hard loss, but the performance of hard loss used in MNLI-m and MNLI-mm is better. All of these experimental groups show that the three different losses used are useful in LRC-BERT.

\subsubsection{Effect of gradient perturbation}

\begin{figure}[t]
	\centering
	\includegraphics[width=0.6\columnwidth]{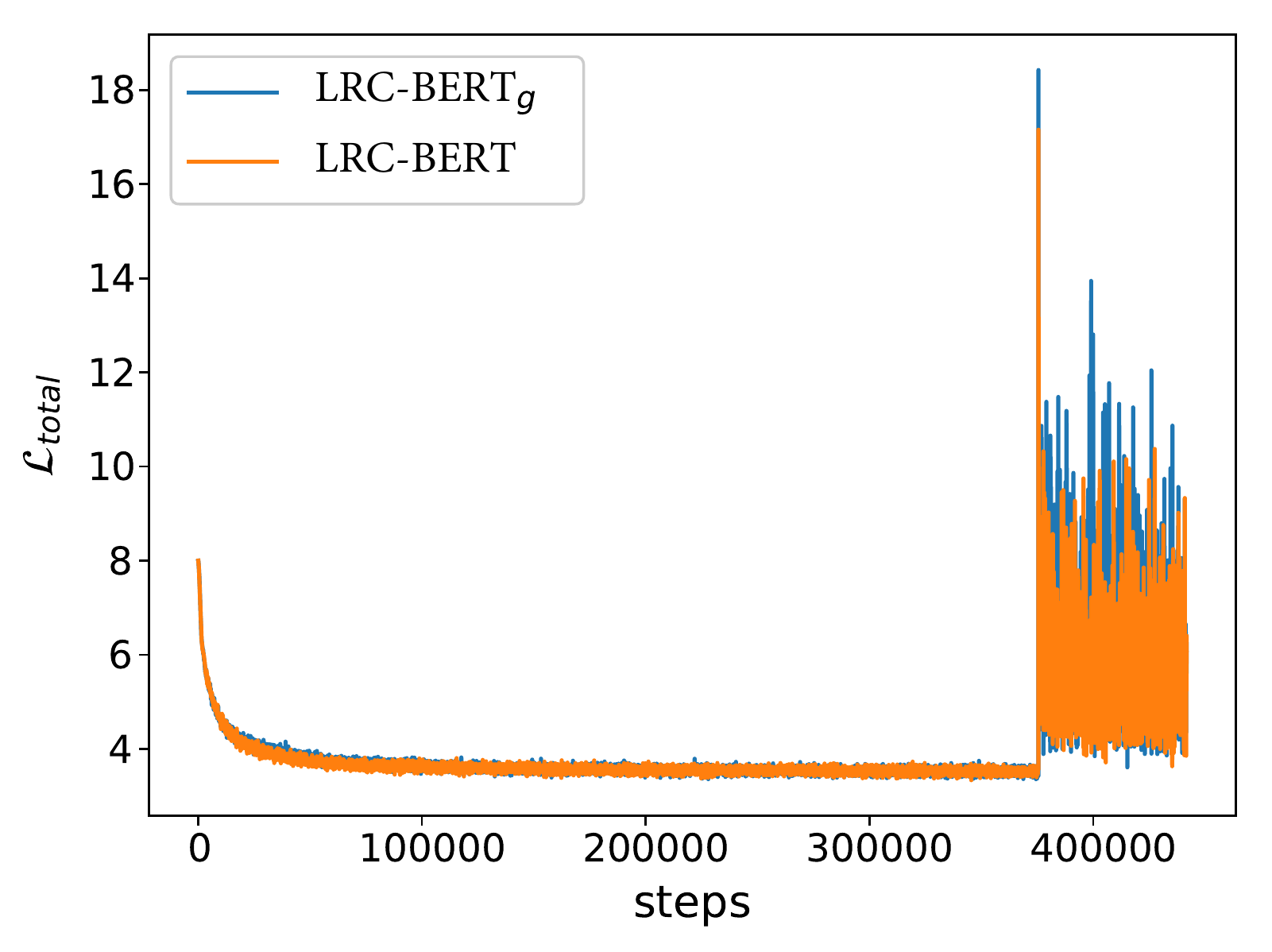} 
	\caption{Effect of gradient disturbance on training loss. The variants are validated on the training set of the MNLI-m task and the loss fluctuation in the second stage of the two-stage training method is obviously increased.}
	\label{fig_5}
\end{figure}

The introduction of gradient perturbation can change the data distribution of the middle layer in the training phase. In order to verify the stability improvement brought by the gradient perturbation on knowledge distillation, we construct LRC-BERT$_g$ after removing the gradient perturbation from the training process. Fig.~\ref{fig_5} shows the training loss of LRC-BERT and LRC-BERT$_g$ on the MNLI-m task. It can be seen from the result that in the first stage of the two-stage training, their downward trend is relatively consistent, but in the second stage, after considering soft loss and hard loss, the training loss of LRC-BERT with gradient perturbation is more stable than LRC-BERT$_g$. It also shows that the gradient perturbation is feasible in knowledge distillation.

\subsection{Analysis of Two-stage Training Method}

The purpose of the two-stage training is to force LRC-BERT to focus only on the distribution characteristics of the data output in the transformer layer of the student model at the beginning of the training. In order to verify its effectiveness, we design LRC-BERT$_2$ without using the two-stage training method. In addition, during the entire training phase of LRC-BERT$_2$, the weighting parameter of the training loss $\mathcal{L}_{total}$ is always set as $\alpha: \beta: \gamma = 1:1:3$. Table~\ref{table-4} shows the evaluation results of LRC-BERT and LRC-BERT$_2$ on the dev set of MNLI-m. Although we can conclude from the ablation studies that COS-NCE based on contrastive learning can effectively transfer the knowledge of the transformer layer in the teacher model to the student, the accuracy of LRC-BERT$_2$ without using the two-stage training method is 4.0\% lower than that of LRC-BERT, which indicates that the knowledge distillation of prediction layer from the beginning will greatly affect the results of contrast distillation in the transformer layer.

\subsection{Analysis of angle-based distribution features}

COS-NCE is proposed to capture the angular similarity feature of the output in the transformer layer, which is not considered by the existing knowledge distillation methods. For this reason, we design a comparison model BERT$_M$, which utilizes MSE for intermediate layer distillation. The average performance of BERT$_M$ is 77.0, which is lower than 77.5 of LRC-BERT. Table~\ref{table-5} shows the four use cases extracted from the dev set of MNLI-m that can be correctly predicted by LRC-BERT, and record the prediction results of BERT$_M$. Specifically, when LRC-BERT and BERT$_M$ predict the first two cases, the MSE and angular distance of the transformer layer are in line with the expected fluctuation range, both can make correct prediction. When the last two cases are predicted, the MSE of BERT$_M$ increased significantly, while angular distance of LRC-BERT has no obvious change. LRC-BERT is trained based on COS-NCE and can still make correct prediction, which shows that the structural characteristics of the output distribution in the middle layer can be effectively used for knowledge distillation. However, in the knowledge distillation method based on Euclidean distance, the prediction result may be completely opposite, which briefly illustrates that the loss function based on angular similarity is feasible in knowledge distillation.

\section{Conclusion}

In this paper, we propose a novel knowledge distillation framework LRC-BERT to compress a large BERT model into a shallow one. Firstly, the COS-NCE based on contrastive learning is proposed to distill the output of the intermediate layer from the angle distance, which is not considered by the existing knowledge distillation methods. Then, we introduce a gradient perturbation-based training architecture in the training phase to increase the robustness of the model, which is the first attempt in knowledge distillation. Finally, to better capture the distribution characteristics of the intermediate layer, we design a two-stage training method for the total distillation loss.  Extensive experiments on 8 datasets provided by GLUE benchmark show that the model is effective.

\bibliography{Formatting-Instructions-LaTeX-2021.bib}

\end{document}